\documentclass[journal]{IEEEtran}
\usepackage{amssymb}
\usepackage{booktabs}
\usepackage{amsmath}
\usepackage{graphicx}
\usepackage{diagbox} % 斜线表格
\usepackage{stfloats}
\usepackage{array} % 表格内容垂直居中
\usepackage{color}
\usepackage{threeparttable}

\usepackage{url}
\usepackage{float}

\usepackage[T1]{fontenc}    % use 8-bit T1 fonts

\begin{document}
\bstctlcite{BSTcontrol}

\newpage
\mbox{This work has been submitted to the IEEE for possible publication. Copyright may be transferred without notice, after}
\mbox{which this version may no longer be accessible.}
\newpage

\title{Graph Polish: A Novel Graph Generation Paradigm for Molecular Optimization}

\author{Chaojie~Ji,
        Yijia~Zheng,
        Ruxin~Wang,
        Yunpeng~Cai,
        and~Hongyan~Wu % <-this % stops a space
\thanks{
Manuscript received XXXX XX, XXXX; revised XXXX XX, XXXX.
% This work was supported in part by the National Natural Science Foundation of China under Grant U1913210, and Grant 61771465, in part by the Strategic Priority CAS Project under Grant XDB38040200, and in part by the Shenzhen Basic Research Projects under Grant JCYJ20180703145002040, and Grant JCYJ20180703145202065).
(Corresponding author: Hongyan Wu.)}
\thanks{Chaojie Ji, Ruxin Wang, Yunpeng Cai and Hongyan Wu are with the Joint Engineering Research Center for Health Big Data Intelligent Analysis Technology, Shenzhen Institutes of Advanced Technology, Chinese Academy of Sciences, Shenzhen, China (e-mail: cj.ji@siat.ac.cn; rx.wang@siat.ac.cn; yp.cai@siat.ac.cn; hy.wu@siat.ac.cn).}% <-this % stops a space
\thanks{Yijia Zheng is with the Shenzhen Institutes of Advanced Technology, Chinese Academy of Sciences, Shenzhen, China, and University of Chinese Academy of Sciences, Beijing, China (e-mail: yj.zheng@siat.ac.cn).}% <-this % stops a space
\thanks{Chaojie Ji and Yijia Zheng make equal contribution.}}% <-this % stops a space

\markboth{IEEE
% TRANSACTIONS ON NEURAL NETWORKS AND LEARNING SYSTEMS
,~Vol.~XX, No.~X, XXXX~XXXX}%
{Shell \MakeLowercase{\textit{et al.}}: Bare Demo of IEEEtran.cls for IEEE Journals}
\maketitle

\begin{abstract}
Molecular optimization, which transforms a given input molecule $X$ into another $Y$ with desirable properties, is essential in molecular drug discovery. The traditional translating approaches, generating the molecular graphs from scratch by adding some substructures piece by piece, prone to error because of the large set of candidate substructures in a large number of steps to the final target. In this study, we present a novel molecular optimization paradigm, Graph Polish, which changes molecular optimization from the traditional ``two-language translating'' task into a ``single-language polishing'' task. The key to this optimization paradigm is to find an optimization center subject to the conditions that the preserved areas around it ought to be maximized and thereafter the removed and added regions should be minimized. We then propose an effective and efficient learning framework T{\&}S polish to capture the long-term dependencies in the optimization steps. The T component automatically identifies and annotates the optimization centers and the preservation, removal and addition of some parts of the molecule, and the S component learns these behaviors and applies these actions to a new molecule. Furthermore, the proposed paradigm can offer an intuitive interpretation for each molecular optimization result. Experiments with multiple optimization tasks are conducted on four benchmark datasets. The proposed T{\&}S polish approach achieves significant advantage over the five state-of-the-art baseline methods on all the tasks. In addition, extensive studies are conducted to validate the effectiveness, explainability and time saving of the novel optimization paradigm.
\end{abstract}


\IEEEpeerreviewmaketitle
\section{Introduction}

\IEEEPARstart{I}{ntroducing} a new drug into the market takes over one billion USD and an average of 13 years \cite{paul2010improve, dimasi2016innovation}.
As part of this task, molecular drug discovery is a critical step in which numerous molecules need to be generated \cite{imrie2020deep}.
% Challenge
This is clearly a formidable task \cite{brown2020artificial}, since the scale of possible drug-like compounds is between $10^{23}$ and $10^{60}$.
% Task Definition
Transforming a given input molecule $X$ into another $Y$ with desirable properties --- the molecular optimization problem --- is essential in molecular drug discovery \cite{griffen2011matched, dossetter2013matched}.

% Generative Methods
Various deep generative models \cite{Moret2020, 9046288, gebauer2019symmetry} have been introduced to generate molecules with specified properties by taking advantage of the representational ability of deep learning methods \cite{ji2020hopgat}.
% RNN-based
By converting molecular graphs into simplified molecular input line system (SMILES) strings \cite{weininger1988smiles, dai2018syntax-directed}, several studies have proposed using recurrent neural networks (RNNs) \cite{8998339} or long short-term memory \cite{ji2020cascade} to generate valid SMILES strings belonging to de novo molecules \cite{gupta2018generative, merk2018novo, arus2020smiles}.
% Autoencoder-based
Variational autoencoders (VAEs) \cite{gomez2018automatic, dai2018syntax, lim2018molecular} and adversarial autoencoders (AAEs) \cite{blaschke2018application, kadurin2017cornucopia, kadurin2017drugan} produce a compound with similar properties to a given compound by sampling the given latent vector with a typical encoder-decoder architecture.
% GAN and RL-based
Generative adversarial networks (GANs) apply two tools, i.e., generative networks and discriminative networks, to play a zero-sum game jointly, and various molecules are generated by the generative networks. Hybrid models are explored in \cite{you2018graph, putin2018reinforced} by combining GANs with reinforcement learning (RL). Combining these generators with a property predictor, the optimization problem can be solved by Bayesian optimization or reinforcement learning \cite{griffiths2020constrained}.
All of these models generate molecules from an input molecule without the direct supervision of explicit target molecules \cite{xue2019advances, grebner2020automated}.

% Turn generative models into supervised learning methods
However, transforming an input molecule into another with optimized properties is more sample-efficient with the guidance of a target through supervised learning \cite{10.1007/978-3-030-47426-3_34}. Recently, supervised graph-to-graph translation has emerged \cite{jin2018learning}. In the variational junction tree encoder-decoder (VJTNN), each molecule can be represented as a junction tree that is composed of multiple substructures, i.e., rings, atoms and bonds. Then, the input molecular graphs and corresponding junction trees are fed into a two-level encoder, and a decoder applies these encoded representations to produce similar compounds to the input that have better properties. Fu et al. further extended the VJTNN with a copy{\&}refine strategy (CORE) in which, in every step, the probability of a candidate substructure being copied is first predicted, and then which substructure is copied, from the source molecule or all possible structures in the training database, is predicted. This method generates a novel target by adding the substructures one by one as in all the other graph translation approaches. \cite{fu2019core}.

% Challenges
% Summarize the common drawbacks of aforementioned methods
Both of the aforementioned methods suffer from the same two challenges: (1) generating the molecular graphs from scratch by adding some substructures piece by piece, which leads to a large number of steps to the final target; (2) the set of all possible substructures is large; e.g., there are approximately 800 unique substructures in the ZINC database \cite{sterling2015zinc}. The prediction error in one of the generation steps could cause the method to exhibit undesirable behaviors. Prior works have uniformly viewed the molecular optimization problem as a full translation problem in which two kinds of language (the source and target graphs) should be aligned and processed word by word.
\begin{figure}[t]
 \centering
 \includegraphics[width=0.49\textwidth]{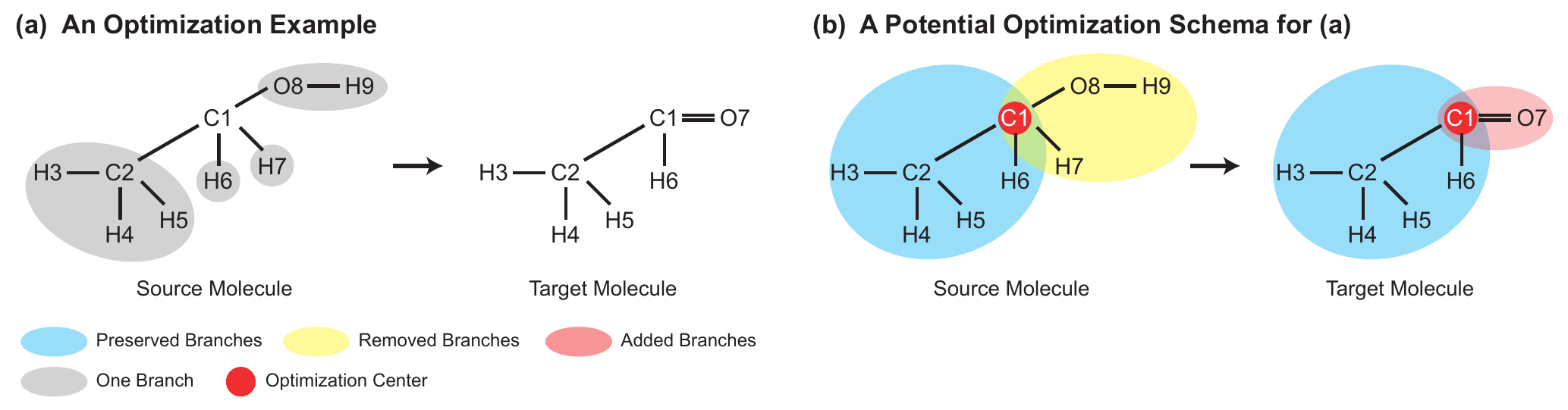}
 \caption{A possible molecular optimization from a source molecule to the target compound. The blue circles represent unchanged areas, and the yellow parts are the domains to be improved. The number beside each atom is a labeled index for readability without chemical meanings.}
 \label{optimization_example}
\end{figure}
However, in most cases, molecular optimization is not a complete interlingual translation activity that proceeds from scratch. Only partial improvement on the input molecules needs to be conducted. For the instance of molecular optimization in Fig. \ref{optimization_example}, the blue circles represent unchanged areas, and the yellow parts are the domains to be improved. Clearly, full translation from scratch directly results in a large substructure search space, and thus, both performance and computational resources may be greatly influenced. Furthermore, explainability, considered a critical factor in deep graph models \cite{ying2019gnnexplainer, baldassarre2019explainability, ji2020perturb}, has been ignored in prior state-of-the-art graph generators.

% Methods
% Contribution 1 - Graph Polishing
Inspired by the above observations, we present a novel molecular optimization paradigm, \textbf{Graph Polish}, which changes molecular optimization from the traditional ``two-language translating'' task into a ``single-language polishing'' task. In this task, the appropriate substructures are first identified and preserved, and then the surrounding context is improved. That is, a molecular optimization operation can be organized as a sequence of actions: given an input molecule, it is first predicted which atom can be viewed as the optimization center, and then the nearby branch regions are optimized around this center. Therefore, the key to the molecular optimization problem is to find an optimization center subject to the conditions that the preserved areas around it ought to be maximized and thereafter the removed and added regions should be minimized. The preserved areas can effectively decrease the number of steps in molecular optimization and guide the subsequent generation of the new substructures as a prior knowledge.

% Contribution 2 - T&S
In addition, we propose an efficient end-to-end model to automatically predict whether we should preserve, delete or add a substructure and how we should generate new substructures. Based on the predefined molecule pairs in a chemistry and drug design knowledge base (e.g., the octanol-water partition coefficient (LogP) database), a framework --- \textbf{Teacher and Student (T{\&}S) polish} --- composed of teacher and student components is proposed to capture long-term dependencies in the optimization steps. All the preferred actions in each step are annotated by the T component, and the S component is responsible for learning the implicit logic and applying the logic to a new molecule.
% Contribution 3 - Explainability
Importantly, these optimization steps naturally offer a reference for researchers to understand the process of molecular optimization.

\begin{figure}[t]
 \centering
 \includegraphics[width=0.49\textwidth]{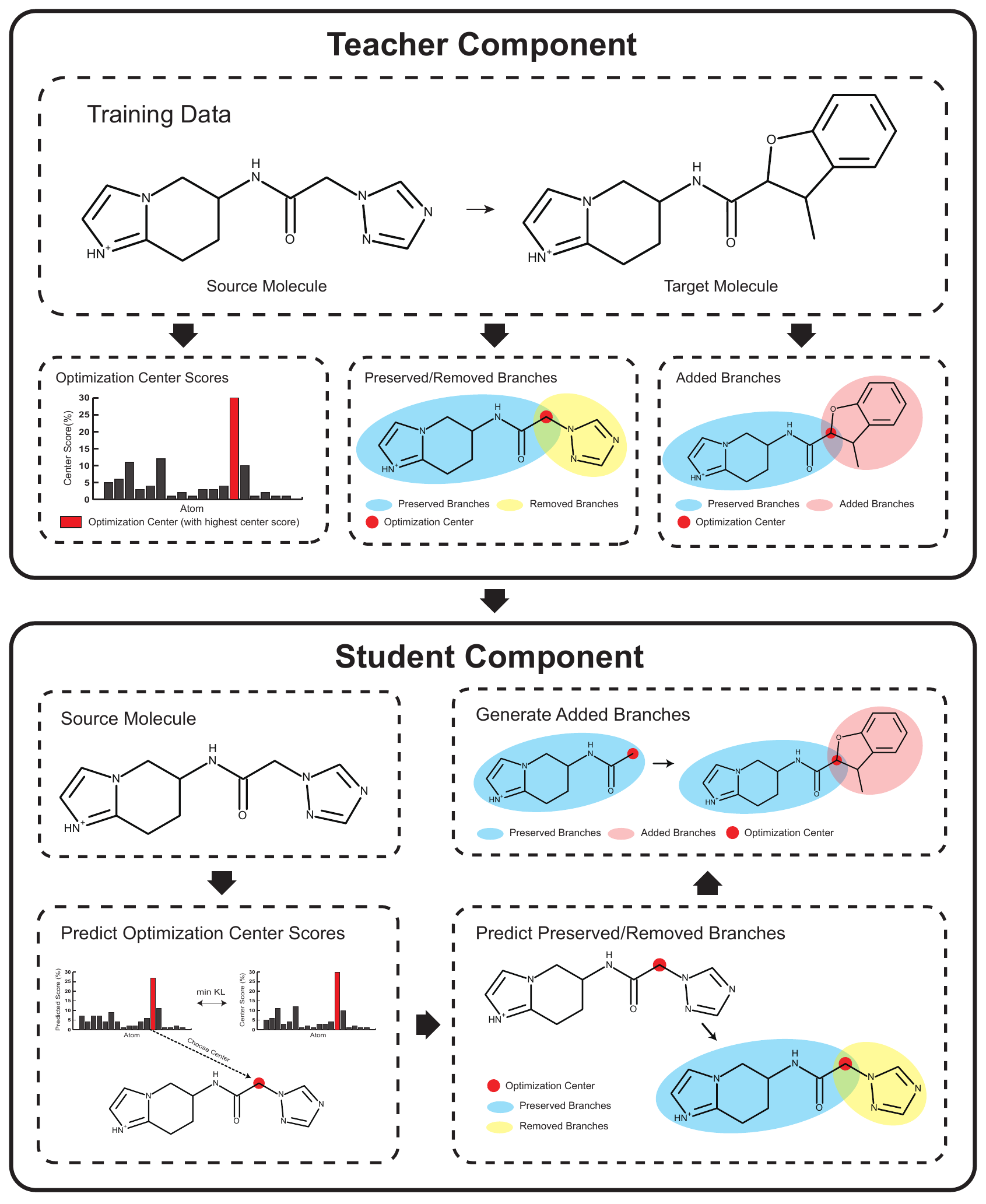}
 \caption{Overall framework of the proposed T{\&}S polish method. The T component automatically identifies the optimization center and the preservation, removal and addition of some parts of a molecule, while the S component learns these behaviors and leverages these actions to create a new molecule.}
 \label{framework}
\end{figure}
X
% Contribution
To summarize, our main contributions are as follows:
\begin{itemize}
  \item To the best of our knowledge, our paper is the first study to propose a novel paradigm, Graph Polish, for molecular property optimization in which optimization centers are first located and polishing actions are performed on surrounding branches. Unlike graph-to-graph translation, the graph polish paradigm can effectively decrease the number of steps of molecular optimization.
  \item An effective and efficient learning framework T{\&}S polish is proposed to capture the long-term dependencies in the optimization steps. As shown in Fig. \ref{framework}, the T component automatically identifies and annotates the optimization centers and the preservation, removal and addition of some parts of the molecule, while the S component learns these behaviors and applies these actions to a new molecule.
  \item An intuitive interpretation for each molecular optimization result is naturally produced by the proposed paradigm. This provides clues for researchers to deepen the understanding of the mechanism for each optimization behavior.
  \item The experiments with multiple optimization tasks are conducted on four benchmark datasets to evaluate the proposed T{\&}S polish approach. Specifically, we assess our model from two perspectives, including the success rate and improved properties. In addition, extensive studies are conducted to validate the effectiveness, explainability and time saving of the novel optimization paradigm.
\end{itemize}

\section{Related Work}
\subsection{Generative Model of Molecules}
Deep generative models for de novo molecule design aim to learn the true data distribution of the compounds in the training set \cite{bradshaw2019model}.
The input can be formulated as a set of pairs $(X, P)$,
where $X$ denotes an input molecule and $P$ is an optional label representing the molecular properties to be optimized \cite{neves2020deep}; then, a new molecule $Y$ with optimized properties are generated \cite{kadurin2017drugan}.
We can categorize these generative approaches in three classes.
% AE-based
The first category is autoencoder (AE)-based methods, in which encoders project high-dimensional input molecules into low-dimensional representations and decoders reconstruct the original inputs according to these low-dimensional representations. Enhanced by additional latent variables and discriminator neural networks, VAE-based \cite{gomez2018automatic, dai2018syntax, lim2018molecular} and AAE-based \cite{blaschke2018application, kadurin2017cornucopia, kadurin2017drugan} approaches are subsequently proposed to generate molecules.
Nevertheless, these AE-based methods intrinsically emphasize feature representation and latent variable modeling, while imperative generation missions are their byproducts \cite{8742794, lim2020scaffold}.
% GAN and RL-based
In contrast, GANs apply two roles \cite{9032341, 9058981}, i.e., generative networks and discriminative networks, to play a zero-sum game jointly; this type of method is explicitly set up for generation tasks. To further constrain the molecular generation process toward desired properties, RL combined with GANs is proposed. Hybrid models are presented \cite{you2018graph, putin2018adversarial}.
% RNN-based
Benefitting from the emergence of SMILES strings, the gap between graphs and natural language is narrowed \cite{bjerrum2018improving}.
Representing molecular graphs as SMILES strings, several studies have applied RNNs to generate valid SMILES strings belonging to de novo molecules \cite{gupta2018generative, merk2018novo, kotsias2020direct}. However, the fatal flaw of SMILES-based methods is that they can produce invalid molecules.

\subsection{Supervised Learning on Molecular Pairs}
By comparison, the concept of matched molecular pairs $\{X, Y\}$, in which a molecule is explicitly transformed into another specified molecule with better properties, is introduced; this is obviously more sample-efficient with the guidance of the target $Y$.
Based on these visible pairs, a supervised learning strategy has been introduced into molecular optimization \cite{jin2018learning}.
Jin et al. the VJTNN with an adversarial component. The main idea is derived from the junction tree variational autoencoder (JTVAE) in which the junction tree is constructed according to multiple chemical substructures, i.e., rings, atoms and bonds \cite{jin2018junction}. The graph and corresponding tree of a molecule are encoded, decoded into a generated scaffolding tree and finally used to produce a new compound.
% CORE
CORE further extends the VJTNN with a copy-refine strategy in which, in every step, the probability of a candidate substructure being copied is first predicted, and then it is predicted which substructure is copied, one from the source molecule or one of all the possible structures in the training database. Although CORE notes that some substructures of the target molecule may come from the source molecule, it still generates a novel target by adding substructures one by one as in all the other graph translation approaches.

Without exception, these approaches all view molecule design as a graph-to-graph translation problem. They generate the target molecules by adding substructures one by one from scratch but ignore that some subgraphs in target molecules are invariant with respect to the input ones, which leads to a limited performance and waste of resources.
% Our work
Our work is closely related to these supervised learning tasks. Concretely, based on the accessible molecule pairs, by maximizing the preserved parts of the source molecule, the proposed T{\&}S polish follows the graph polish paradigm and automatically generates a relatively small part of the target molecule.

\section{Preliminaries and Problem Formulation}
% General Concepts in Graphs (Molecules)
Given a molecule, we represent it as $G=(V_G, E_G)$, where $V_G$ and $E_G$ are the sets of atoms and bonds inside the molecule, respectively. In addition, $|V_G|$ stands for the number of elements (nodes) in the set $V_G$. $x_i$ indicates the feature vector of node $i$, and $x_{i,j}$ is the type of bond between nodes $i$ and $j$.
% Mission Target
Formally, our method is designed to generate a target molecular graph $Y=(V_Y, E_Y)$ according to the given source molecular graph $X=(V_X, E_X)$.

% Extension Concepts in Graphs
We now introduce several concepts that are applied later in our paper. $N(i)$ is a set used to symbolize the neighbor nodes around node $i$.
We consider that a graph $G$ can be split into a set of subgraphs by cutting all the edges around a given node $i$, as shown in the gray areas in Fig. \ref{optimization_example}(a). For a node $j \in N(i)$, a branch $ b^G_{i,j}$ is an area that starts from node $j$ around node $i$ with respect to graph $G$. The four gray areas in Fig. \ref{optimization_example}(a) correspond to the branches $b^G_{C1,C2}$, $b^G_{C1 ,H6}$, $b^G_{C1,H7}$ and $b^G_{C1,O8}$. $B^G_{i}$ is the set composed of all the branches around node $i$ with respect to graph $G$. $B^G_{C1}$ comprises the four gray areas.
Each branch is also a graph, denoted as $b^G_{i,j}=(V_{b_{i,j}}, E_{b_{i,j}})$. In our paper, graphs $G_1$ and $G_2$ are represented as isomorphic by writing either $G_1 = I(G_2)$ or $G_2 = I(G_1)$.

\section{Method}

\begin{table}[!htbp]
  \caption{Important notations used in this paper}
  \centering
  \begin{tabular}{c|l}
    \hline
    \multicolumn{1}{c|}{Notations} & Short explanation \\
    \hline
      $te$ & the teacher component \\
      $\varepsilon^{te}_r$ & signal of a branch to be preserved (r) \\
      % $\epsilon^{oc}$ & score of a node as an optimization center  \\
      $s^{te}_i$ & score of node $i$ to be an optimization center \\
      $c^{te}$ & preferred optimization center in the source molecule \\
      $c^{te'}$ & the mapped node ($'$) in the target molecule of \\
      &  the optimization center $c^{te}$ \\
      $U^{te}_r$ & the set of preserved (r) branches \\
      % $\epsilon^{ad}$ & signal of a branch to be added \\
      $U^{te}_{ad}$ & the set of branches to be added (ad) \\
      \hline
      $st$ & the student component \\
      $s^{st}_i$ & score of node $i$ to be an optimization center \\
      $c^{st}$ & predicted optimization center in source molecule \\
      $\varepsilon^{st}_r$ & signal of a branch to be preserved (r) \\
      $U^{st}_r$ & the set of preserved (r) branches \\
      $H^*$ & encoded node representations of a graph or tree (*) \\
      $q_t$ & distribution of topological prediction at the $t$-th step \\
      $p_t$ & distribution of label prediction at the $t$-th step \\
      $f^s(\cdot)$ & score (s) function to select the decoded molecules \\
    \hline
  \end{tabular}
  \label{notations}
\end{table}

Fig. \ref{optimization_example} shows a standard molecular optimization process in which a source molecule is converted into a target compound. To perform the molecular optimization automatically, it is essential to locate an optimization center and identify the candidate optimization actions on each branch around the center. A framework T{\&}S polish is proposed to capture long-term action dependencies during the optimization steps based on the predefined molecule pairs in the chemistry and drug design knowledge base. All preferred actions in each step are annotated by the T component, and the S component is responsible for learning the latent logic and leveraging the logic to create a new molecule. Table \ref{notations} lists some important notations that will be used in the rest of the paper.

\subsection{Teacher Component}
% Intuition of TG
The task of the T component is to monitor a large number of molecule pairs composed of source and target molecules in the chemistry and drug design knowledge base and to identify and annotate the possible action patterns for each pair.

\subsubsection{Action Pattern}

% Two Aims
% For First Aim, Design 3 Actions
The key to molecular optimization is to find a center subject to the conditions that the preserved areas around it are maximized and the removed and added regions are minimized. Thus, identifying optimization centers is considered a meta-action. We design three more meta-actions --- preservation, removal and addition operations --- on the branches around a node in a source graph. Concretely, based on a given node, preserving a branch means that the branch will remain unchanged in the target graph, while removing it indicates that the branch will no longer occurs in the target graph. Last, the branches not belonging to source but appearing in the target are labeled as branches to be added later, as illustrated in Fig. \ref{optimization_example}. Once these actions have been performed sequentially on the source molecule, the target molecule can be produced precisely.

\subsubsection{Action Identification}
In this section, we detail the four meta actions and assemble them into a whole procedure that can be applied to any molecule pairs.

\textbf{$\bullet$ \ Selecting Preserved and Removed Branches}

An optimization center is determined according to whether minimal changes around it are caused during the optimization procedure. Minimizing changes around the optimization center is needed to maximize the scale of the preservation area. Therefore, we first show how to select the preserved and removed branches.

The branch $b^X_{i,k}$ to be preserved in the source molecule $X$ can be projected to the corresponding isomorphic subgraph $b^Y_{j,l}$ in the target $Y$, e.g., $b^X_{C1,C2}$ and $b^Y_{C1,C2}$ in Fig. \ref{optimization_example}(b).
In addition, removal is an alternative to preservation. That is, a branch in $X$ must be either preserved or removed.

We assign $\varepsilon^{te}_r \in \{0, 1\}$ to represent the branch $b^X_{i,k}$ to be preserved or removed:

\begin{equation}
  \label{Method:Selecting Reserved Branches:E1}
  q^{r}(\varepsilon^{te}_r|i,k,j)=
  \begin{cases}
    1, & \text{if}\ b^X_{i,k}=I(b^Y_{j,l}), \forall b^Y_{j,l}\in B^Y_j \\
    0, & \text{otherwise}
  \end{cases}
\end{equation}
where $i$ is a candidate optimization center in $X$ and $j$ is a candidate node in $Y$ mapped from node $i$. The mapping and designation strategies, from node $i$ to $j$, will be described in the section of locating optimization center. In addition, $k$ and $l$ indicate the starting node of a branch in the source molecule around node $i$ and its corresponding mapped branch in the target molecule around node $j$, respectively. Here, we should note that the probability of these two actions greatly depends on the given starting nodes as well as the optimization center.

% Special Case
In some cases, there are multiple identical branches in $X$ and only one isomorphic subgraph in $Y$. In such a situation, only one branch, not all of them, in $X$ should be preserved. To implement this, when $q^{r}(\varepsilon^{te}_r|i,k,j)=1$, we exclude the corresponding $b^Y_{j,l}$ as follows:
\begin{equation}
  \label{Method:Selecting Reserved Branches:E2}
  B^Y_j = B^Y_j - b^Y_{j,l}
\end{equation}

This operation is iteratively performed on all preserved branches for nodes $i$ and $j$, which affects the execution of Equation (\ref{Method:Selecting Reserved Branches:E1}) for all remaining branches in the next iteration.

\textbf{$\bullet$ \ Locating Optimization Center}

Optimization centers highlight the most ``stable'' vertices, in which the size of preserved subgraphs is the largest. We define the size as the total number of atoms inside all preserved subgraphs.

In particular, optimization centers are pairs such that the node $i$ resides in the source graph and its corresponding partner is the mapped node $j$ in the target, e.g., the atom pair $(C1, C1)$ in the source and target graph in Fig. \ref{optimization_example}(b). We define the score of a candidate optimization center $i$ in $X$ with a candidate mapped node $j$ in $Y$ as $\varepsilon^{c}$:

\begin{equation}
  \label{Method:Locating Optimization center:E1}
  q^{c}(\varepsilon^{c}|i,j) = \sum_{b^X_{i,k}\in B^X_i}|V_{b^X_{i,k}}|q^{r}(\varepsilon^{te}_r|i,k,j)
\end{equation}

The search space from $i$ to mapped nodes $j$ could be very large and pose a challenge to finding the best center pair.
% Partial Solution
 The first constraint applied in this study is that the chemical elements for the members of the pair should be identical.
We then adopt a greedy search strategy, traversing all the distilled candidate mapped nodes in the first step and picking out the atoms with the maximum score from node $i$.
\begin{equation}
  \label{Method:Locating Optimization center:E3}
  s_i=\mathop{\max} \{q^{c}(\varepsilon^{c}|i,j)|i = j, \forall j\in V_Y\}
\end{equation}
where $i = j$ denotes that nodes $i$ and $j$ are identical chemical elements.

To make the maximum scores easily comparable across different candidate optimization centers, we normalize them across all candidates of $i$ using the softmax function:
\begin{equation}
  \label{Method:Locating Optimization center:E2}
  s^{te}_{i}=\frac{exp(s_i)}{\sum_{k\in V_X}exp(s_k)}
\end{equation}

Finally, the chosen optimization center $c^{te}$ and mapped node ${c^{te}}'$ can be identified by:
\begin{equation}
  \label{Method:Locating Optimization center:E4}
  c^{te}=\mathop{\arg\max}_i \{s^{te}_i|i\in V_X\}
\end{equation}
\begin{equation}
  \label{Method:Locating Optimization center:E5}
  {c^{te}}' = \mathop{\arg\max}_j \{q^{c}(\varepsilon^{c}|c^{te},j)|c^{te} = j, \forall j\in V_Y\}
\end{equation}

% \begin{equation}
% \begin{split}
%   \label{Method:Locating Optimization center:E5}
%   {c^{te}}' = \mathop{\arg\max}_j \{&q^{c}(\varepsilon^{c}|c^{te},j)\\
%   & |c^{te} = j, \forall j\in V_Y\}
% \end{split}
% \end{equation}

Once optimization centers $c^{te}$ and ${c^{te}}'$ are obtained as described above, the set of preserved branches $U^{te}_r$ is simultaneously obtained, which will be used in the next section:
\begin{equation}
  \label{Method:Locating Optimization center:E7}
  U^{te}_r=\{b^X_{c^{te},k} | q^{r}(\varepsilon^{te}_r|c^{te},k, {c^{te}}')=1, \forall k\in N(c^{te})\}
\end{equation}

\textbf{$\bullet$ \ Choosing Added Branches}

The branches around ${c^{te}}'$ not belonging to $U^{te}_r$ can be regarded as the added branches:
\begin{equation}
  \label{Method:Selecting Added Branches:E1}
  q^{ad}(\varepsilon^{ad}|{c^{te}}',k)=
  \begin{cases}
  1, & \text{if}\ \forall I(b^Y_{{c^{te}}',k})\notin U^{te}_r \\
  0, & \text{otherwise} \\
  \end{cases}
\end{equation}

Similar to the motivation of Equation (\ref{Method:Selecting Reserved Branches:E2}), multiple identical branches in $Y$ correspond to one isomorphic subgraph in $X$. Thus, we also iteratively remove $b^Y_{{c^{te}}',k}$ from $U^{te}_r$ when $q^{ad}(\varepsilon^{ad}|{c^{te}}',k)=0$, as follows:

\begin{equation}
  U^{te}_r = U^{te}_r - I(b^Y_{{c^{te}}',k})
\end{equation}
which is executed iteratively, in the same way as selecting preserved and removed branches.

Please note that the added branches are annotated with respect to the target molecule. We collect these added branches into a set:
\begin{equation}
  \label{Method:Selecting Added Branches:E2}
  U^{te}_{ad} = \{b^Y_{{c^{te}}',k}|q^{ad}(\varepsilon^{ad}|{c^{te}}',k)=1,\forall k\in N({c^{te}}')\}
\end{equation}
which are considered target subgraphs to be generated later in the S component.

\subsection{Student Component}
% Necessity of introducing SF
The T component identifies and annotates the action patterns with the supervision of the source and target molecule pairs. By comparing the difference between the source and target molecules, the T component provides a reasonable solution to reproduce the optimization process. However, the procedure cannot be directly extended to the S component since the target molecules are unseen and the difference between the source and target molecules is inaccessible. The S component takes charge of optimizing a novel molecule without any guidance from the target molecules. In this section, we propose a solution that helps the S component study the logic implied in the T component in the training session.
Furthermore, we reorganize the solution procedure composed of choosing the preserved, removed and added branches and seeking optimization centers to enable the S component to have self-judgment at testing stage.

\subsubsection{Predicting Optimization Center}
\label{Predicting_Optimization_center}
Although we cannot determine the optimization centers by comparing the source and target molecules in the S component, the distribution of candidate optimization center scores is available in the T component. The S component can learn the same logic as the T component by minimizing the Kullback-Leibler divergence function between the S and T components:
\begin{equation}
  \label{Method:Predicting Optimization center:E6}
  \mathcal{L}^{c}=\sum_{k\in V_X}D_{KL}(s^{te}_k||s^{st}_k)
\end{equation}
where $s^{te}_k$ and $s^{st}_k$ are the distributions of the candidate optimization center score in the T and S components, respectively.

To represent $s^{st}_k$, we apply graph message-passing networks (MPNs) \cite{10.5555/3045390.3045675, gilmer2020message} to encode each molecule as a graph. For each single node (atom), the networks update the node representation $h_i$ by aggregating the neighbor-updated messages $m^t_{i,j}$ that are delivered from node $i$ to $j$ in the $t$-th iteration with edge representation $x_{i,j}$.

\begin{equation}
  \label{Method:Predicting Optimization center:E1}
  m^t_{i,j}=f_1(x_i,x_{i,j},\sum_{k\in N(i)\backslash j}m^{t-1}_{k,i})
\end{equation}

After $\mathcal I$ iteration steps, we can obtain $h_i$ as:
\begin{equation}
  \label{Method:Predicting Optimization center:E2}
  h_i=f_2(x_i, \sum_{k\in N(i)}m^\mathcal I_{k,i})
\end{equation}
where $m^0_{i,j}$ is initialized as $0$. $f_*(\cdot)$ symbolizes an independent neural network.

To reduce the overfitting caused by an abundance of parameters, we share the node representation across all prediction subtasks in the S module. Once the node representation is fixed, the entire source molecule can be encoded as:
\begin{equation}
  \label{Method:Predicting Optimization center:E3}
  h_X=\sum_{k\in V_X}\frac{h_k}{|V_X|}
\end{equation}

As mentioned in the discussion of the T component, an optimal optimization center minimizes the changes from the source to the target molecule. Thus, both the candidate optimization center and the entire molecular structure deserve attention. Therefore, we formally predict the likelihood of an atom $i$ being an optimization center as:
\begin{equation}
  \label{Method:Predicting Optimization center:E4}
  s_i=f_3([h_X, h_i])
\end{equation}
where $[\cdot]$ indicates vector concatenation.
For easier comparison across all candidate optimization centers and alignment with $s^{te}_i$, we further normalize $s_i$:
\begin{equation}
  \label{Method:Predicting Optimization center:E5}
  s^{st}_i=\frac{exp(s_i)}{\sum_{k\in V_X}exp(s_k)}
\end{equation}

The neural network is trained by minimizing the loss function in Equation (\ref{Method:Predicting Optimization center:E6}). Finally, the optimization center $c^{st}$ can be identified as:
\begin{equation}
  \label{Method:Predicting Optimization center:E7}
  c^{st}=\mathop{\arg\max}_k \{s^{st}_k|k\in V_X\}
\end{equation}

\subsubsection{Predicting Preserved and Removed Branches}
The representation of each branch must be introduced since the subject of removal and preservation is a branch:
\begin{equation}
  \label{Method:Predicting Removed and Reserved Branches:E1}
  h_{b^X_{c^{st},j}}=\sum_{k\in b^X_{c^{st},j}} \frac{h_k}{|V_{b^X_{c^{st},j}}|}
\end{equation}
where $b^X_{c^{st},j}$ is a branch in which $c^{st}$ is the candidate optimization center and j is the starting vertex of this branch around vertex $c^{st}$.

Just as in the T component, preservation is an alternative to removal in S. We also designate a signal $\varepsilon^{st}_r\in \{0, 1\}$, which is drawn from a Bernoulli distribution:
\begin{equation}
  \label{Method:Predicting Removed and Reserved Branches:E2}
  p(\varepsilon^{st}_r|{c^{st},j})=\sigma(f_4([h_{c^{st}},h_{b^X_{c^{st},j}},h_{U^{st}_{t-1}}]))
\end{equation}
where $j$ is the starting node of the branch around node $c^{st}$. $h_{c^{st}}$, $h_{b^X_{c^{st},j}}$ and $h_{U^{st}_{t-1}}$ represent the representation of the optimization center, currently predicted branch and union of already preserved branches, respectively.

The previously preserved branches usually affect the judgment regarding the next branches.
In this study, a branch is classified into the preserved or removed subgraph in each step according to sequence of the SMILES representation of the source molecule. Once a branch is ascertained to be preserved, it will be collected into a set of already-preserved branches:

\begin{equation}
  \label{Method:Predicting Removed and Reserved Branches:E3}
  U^{st}_t=
  \begin{cases}
    U^{st}_{t-1}\cup \{b^X_{c^{st},j}\}, & \text{if}\  p(\varepsilon^{st}_r|c^{st},j)\geq 0.5, t>0 \\
    U^{st}_{t-1}, & \text{if}\ p(\varepsilon^{st}_r|c^{st},j) < 0.5, t>0 \\
    \varnothing, & \text{otherwise}
  \end{cases}
\end{equation}
in which $U^{st}_{t-1}$ represents the set of the last preserved branches in the $t$-th iteration.

We employ the following representation of previously-preserved subgraphs:
\begin{equation}
  \label{Method:Predicting Removed and Reserved Branches:E4}
  h_{U^{st}_{t}}=\sum_{b\in U^{st}_{t}}\frac{h_b}{|U^{st}_{t}|}
\end{equation}
where $b$ indicates one of the subgraphs in the set $U^{st}_{t}$. Clearly, this iteratively accumulated graph set $U^{st}_{t}$ affects the calculation of Equation (\ref{Method:Predicting Removed and Reserved Branches:E2}) for next iteration $t+1$.

For simplicity, we will use $U^{st}_r$ to represent the final preserved branches after $|N(c^{st})|$ iterations, the number of neighbor nodes around optimization center $c^{st}$. The $U$ notation without specially specifying iteration numbers means the final iteration result in this paper.

To make the model learn the distribution of the preserved branches in the T component, we use the following cross-entropy loss function:
\begin{equation}
  \label{Method:Predicting Removed and Reserved Branches:E5}
  \begin{split}
  \mathcal L^{r} = -\sum_{j\in N(c^{te})} q^{r}(\varepsilon^{te}_r|c^{te},j,{c^{te'}})log(p(\varepsilon^{st}_r|c^{te},j))
  \end{split}
\end{equation}
where $\varepsilon^{te}_r$ is the distribution of the preserved branches in the T component.
Specifically, at the training stage, we apply teacher forcing by feeding the predictor the ground-truth optimization center $c^{te}$ and mapped node $c^{te'}$ as input.
During testing, we directly infer the actions on branches around $c^{st}$ without the ground truth $c^{te}$ and $c^{te'}$.

\subsubsection{Predicting Added Branches}
According to obtained optimization center, preserved and removed branches in a source molecule, the remaining steps of generating a new molecule include reformulating and generating addition branches.

\textbf{$\bullet$ \ Reformulating Added Branches}

In a typical graph-to-graph translation task, the molecule generator can be formulated as a model $p(Y|X)$. This model suffers from two drawbacks: (1) substructures are added one by one from scratch through a large number of steps; and (2) the size of the candidate substructure set in each step is approximately 800, which severely affects the performance. To narrow the gap between X and Y, the graph polish paradigm simultaneously attaches the raw ``questions'' (X) with more heuristic tips and simplifies the original ``answers'' (Y) by excluding the known parts, which can greatly improve the rate of obtaining correct answers.

%(1) Attach Questions (X) with More heuristic Tips

%provide an effective clue. It's that the reserved subgraphs
% Intuition
As shown in Equation (\ref{Method:Predicting Removed and Reserved Branches:E3}), the established preservation branches record the invariant parts in the optimization process and offer a tip rule for the generation of the remaining variant parts. All of these branches share a common atom --- the optimization center. To emphasize the relationships among these branches, we merge them into an integrated graph $R=(V_{R}, E_{R})$, in which:
\begin{equation}
  \label{Method:Predicting Added Branches:E1}
  V_{R}=\cup_{b\in U}V_b
\end{equation}
\begin{equation}
  \label{Method:Predicting Added Branches:E2}
  E_{R}=\cup_{b\in U}E_b
\end{equation}
where $U = U^{te}_r$ at the training stage when the technique of teacher forcing is applied, while $U = U^{st}_r$ in the testing phase. Now, the input used in our generator is changed from $(X)$ to $(X, R)$, which is a combination of the original source molecule and the merged preservation branches. Intuitively, $R$ can be considered from the local perspective that partially established elements are highlighted, while $X$ is the global outlook in which the complete input is observed.

%(2) Distill Unknown Parts from Answers (Y)

The raw target molecule can be split into two groups --- the preserved and added areas --- and now the preserved subgraphs are known. Therefore, we can convert the ultimate output from the entire target molecule to the smaller unknown part --- the branches to be added.

We also integrate all elements in $U^{te}_{ad}$ to form an synthesis graph to be added $A=(V_{A}, E_{A})$:
\begin{equation}
  \label{Method:Predicting Added Branches:E3}
  V_A=\cup_{b\in U^{te}_{ad}}V_b
\end{equation}
\begin{equation}
  \label{Method:Predicting Added Branches:E4}
  E_A=\cup_{b\in U^{te}_{ad}}E_b
\end{equation}

Now, we can formulate the subsequent molecular optimization as $p^{ad}(A|X, R)$.

%\textbf{Answer the Questions}

\textbf{$\bullet$ \ Generating Added Branches}

\begin{figure}[t]
 \centering
 \includegraphics[width=0.49\textwidth]{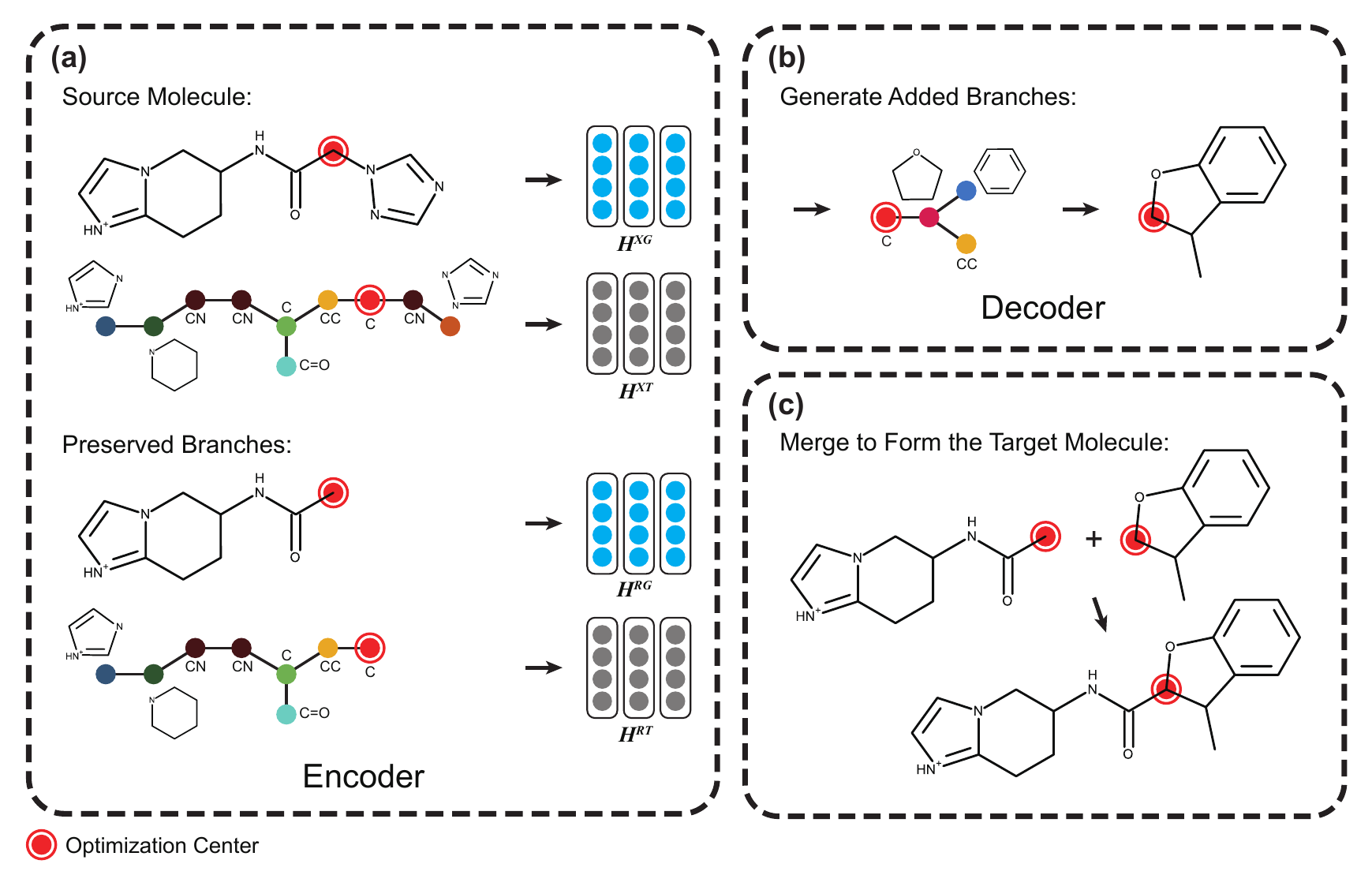}
 \caption{An overview of our method for generating the additional branches and combining them with the preserved area to obtain the target molecule. (a) During the encoder phase, both the source molecule and the preserved subgraph, with their corresponding junction trees, are encoded by two MPNs. (b) At the decoding stage, the above embedded representations are first applied to produce a junction tree, and then the junction tree is further decoded into a molecular subgraph (a subgraph in the target molecule). (c) Finally, the final target molecule is determined, and then the decoded-molecule subgraph and previously preserved branches are merged via the optimization center.}
 \label{branches_generate_overview}
\end{figure}

This section shows how to generate the remaining parts of the target molecule $A$, once the source molecule $X$ is given and the preserved region $R$ is identified, which is illustrated in Fig. \ref{branches_generate_overview}. Our proposed generation method has a similar foundation as the graph-to-graph method \cite{jin2018learning}, but we integrate the additional input information under the guidance of the preserved branches. We emphasize the differences that our proposed method introduces in this section.

(1) Encoder:
% Definition of junction tree
A junction tree is a tree-based representation of a molecule consisting of many subgraph components with valid chemical structures, such as rings and edges, which helps avoid invalid results in molecule generation. Fig. \ref{branches_generate_overview}(a) shows a molecular graph and its corresponding junction tree.
% Main difference among our methods and others
Benefitting from our proposed graph polish optimization paradigm, the preserved branches (graph R) identified previously, together with the source molecule (graph X), are simultaneously fed into the generator, while other methods can only access the source molecule. Consequently, these two parts need to be encoded via our encoders.

% Graph Encoder
First, we define a graph encoder to represent a molecular graph (subgraph).
We use a MPN with a similar architecture as Equations (\ref{Method:Predicting Optimization center:E1}) and (\ref{Method:Predicting Optimization center:E2}) to encode the two graphs. The MPN is shared between the encoding of the source molecule and the preserved subgraph. Accordingly, the final encoded node features of these two graphs are respectively denoted as two sets, i.e., $H^{XG}=\{h^{XG}_1, h^{XG}_2, \cdots \}$ and $H^{RG}=\{ h^{RG}_1, h^{RG}_2, \cdots \}$.
% Tree Encoder
Then, another MPN is utilized to encode the molecular graphs as junction trees (tree-shaped graphs), and the encoded vertex sets of junction trees are symbolized as $H^{XT}=\{ h^{XT}_1, h^{XT}_2, \cdots \}$ and $H^{RT}=\{ h^{RT}_1, h^{RT}_2, \cdots \}$.

(2) Decoder:
The entire decoding procedure comprises two phases: the tree decoder and graph decoder. The tree decoder translates the encoded representations into a junction tree. Then, the graph decoder is employed to produce a specific molecule according to this decoded junction tree.

\textbf{Tree Decoder}
% Overall Mission Definition
The aim of the tree decoder is to assemble a scaffold tree node by node in a depth-first walk.
To be more concrete, the generator starts from a root node and recursively decides whether to expand a child node or backtrack to the parent node at each time step; this is called topological prediction. In particular, following an expansion decision, a specific structure of the expanded child node should be chosen from a predefined component dictionary, which is termed label prediction. Importantly, the moment when backtracking returns to the root node is a signal to terminate the entire generation.

We choose the optimization center as the root node because of its strong stability, where the most preserved branches surround it and all branches are connected via it.

% Search process
Starting from the optimization center, the new branches are recurrently predicted. A tree-based recurrent neural network is applied to record the intermediate status at every time step. Specifically, the edge set, i.e., $\tilde{\cal E} = \{(i_1, j_1), (i_2, j_2), \cdots, (i_m, j_m) \}$, is recorded with a depth-first traversal, and each edge must be traversed twice in both directions--from the top down and then backtracking from bottom to top. The message vector $h_{i_t, j_t}$ at the $t$-th step can be calculated by:

\begin{equation}
  \label{Method:Predicting Added Branches:E7}
  h_{i_t, j_t} = GRU(x_{i_t}, \{ h_{k, i_t} \}_{(k ,i_t) \in \tilde{\cal E}_t, k \neq j_t})
\end{equation}
where $\tilde{\cal E}_t$ is the set of the first $t$ edges in $\tilde{\cal E}$ and $GRU(\cdot)$ stands for a tree gated recurrent unit \cite{jin2018junction}.

With the updated message vector $h_{i_t, j_t}$ and the previously encoded information $H^{XG}$, $H^{RG}$, $H^{XT}$ and $H^{RT}$, the tasks of topological and label prediction can be executed.

a) Topological Prediction: At each time step, for the current node $i_t$, topological prediction aims to decide either to expand a new child node or backtrack to the parent node, which is influenced by the node feature $x_{i_t}$, the inward message $h_{k, i_t}$, and the previously encoded information $H^{XG}$, $H^{RG}$, $H^{XT}$ and $H^{RT}$. This task can thus be considered binary prediction, as follows:
\begin{equation}
  \label{Method:Predicting Added Branches:E8}
  h_t = \tau (f_5(x_{i_t}) + f_6(\sum_{(k, i_t) \in \tilde{\cal E}}h_{k, i_t}))
\end{equation}

\begin{equation}
  \label{Method:Predicting Added Branches:E9}
  c_t^d = [a_1(H^{XG}), a_1(H^{RG}), a_1(H^{XT}), a_1(H^{RT})]
\end{equation}

%   c_t^d = \mathop{concate}_{H\in \{H^{XG}, H^{RG}, H^{XT}, H^{RT}\}} a_1(h_t, H)

\begin{equation}
  \label{Method:Predicting Added Branches:E10}
  p_t = \sigma(u^d \cdot \tau (f_7(h_t) + f_8(c_t^d)))
\end{equation}
where $u^d$ is a trainable parameter, $\tau(\cdot)$ is the ReLU function \cite{nair2010rectified} and $a_1(\cdot)$ calculates the attention scores across all vectors inside $H^*$ and obtain overall representation of $H^*$. $\cdot$ represents a dot-product function. Specifically, this function can be formulated as:
\begin{equation}
  [\alpha_1, ...] = Softmax([f_9(h_t) \cdot h^*_1, ...])
\end{equation}
\begin{equation}
  a_1(H^*) = \sum_{i=1}^{|H^*|} \alpha_i h^*_i
\end{equation}
where $h^*_i$ indicates the $i$-th element in the set $H^*$ and $|H^*|$ is the number of elements in $H^*$.

b) Label Prediction: Once a new child is determined to be expanded, label prediction is conducted to decide which structure will be attached. We predict $q_t$, the distribution over the component dictionary collected from the knowledge bases:
\begin{equation}
  [\alpha_1, ...] = Softmax([f_{10}(h_{i_t, j_t}) \cdot h^*_1, ...])
\end{equation}
\begin{equation}
  a_2(H^*) = \sum_{i=1}^{|H^*|} \alpha_i h^*_i
\end{equation}
\begin{equation}
  \label{Method:Predicting Added Branches:E11}
  c_t^l = [a_2(H^{XG}), a_2(H^{RG}), a_2(H^{XT}), a_2(H^{RT})]
\end{equation}
\begin{equation}
  \label{Method:Predicting Added Branches:E12}
  q_t = Softmax(u^l \cdot \tau (f_{11}(h_{i_t, j_t}) + f_{12}(c_t^l)))
\end{equation}
where $u^l$ are trainable parameters.

\textbf{Graph Decoder}
\begin{figure}[t]
   \centering
   \includegraphics[scale=1.0]{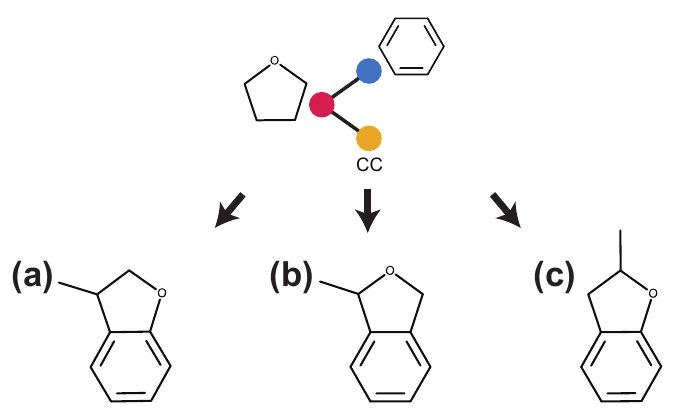}
   \caption{A junction tree can be translated into multiple molecules according to distinct assembly methods between neighbors.}
   \label{assemble_junction_tree}
\end{figure}
% problem definition
After tree decoding is performed, we obtain a junction tree. It is noteworthy that a single junction tree may correspond to multiple molecules, as shown in Fig. \ref{assemble_junction_tree}. All possible candidate attachments at tree node $i$ in the junction tree are enumerated to finally construct a candidate graph set $\mathcal{G}_i = \{ G_{i,1}, ...\}$. Then, a graph decoder is designed to select the correct attachment from $\mathcal{G}_i$.
% Solution
We first apply an MPN over graph $G_{i,j}$ to calculate node representations $H^{DG_{i,j}}$ as the calculation of $H^{XG}$.
Afterwards a scoring function $f^s(\cdot)$ is used to rank all candidates within the set $\mathcal{G}_i$. With our proposed polishing strategy, both the source molecule and the preserved part can be used to guide graph decoding:
\begin{equation}
  f^s(G_{i,j}) = \sum_{k\in V_{G_{i,j}}}h^{DG_{i,j}}_k \cdot f_{13}([\sum_{h\in H^{XG}} h, \sum_{m\in H^{RG}} m])
\end{equation}

Subsequently, a loss function is defined to maximize the likelihood of ground-truth subgraphs $G_i^* \in \mathcal{G}_i$ at all tree nodes $i$.
\begin{equation}
  \label{Method:Predicting Added Branches:E13}
  \mathcal{L}_a = \sum_{i} \left(f^s(G_i^*) - \log \sum_{G'_{i,j} \in \mathcal{G}_i} \exp(f^s(G'_{i,j}))\right)
\end{equation}

Finally, we merge the preserved area and the generated part via the optimization center, polishing and resulting optimized molecule.

\subsection{Characteristics of the T{\&}S Polish}

\textbf{$\bullet$ \ Property 1: Strong Error Tolerance}

Benefiting from the proposed graph generation paradigm and T\&G polish framework, the error rates of molecular optimization can be greatly decreased because (1) the target molecules are generated with guidance from both the preserved part and the source graph and (2) the size of new parts that need to be produced are decreased.

\textbf{$\bullet$ \ Property 2: Low Computational Complexity}

We generate target molecules from one source in a local manner instead of in a global way, as other approaches do. The T{\&}S polish preserves the existing branches to the largest extent, generates a minimal subgraph and attaches them together, which tremendously reduces the computational resources needed.

\textbf{$\bullet$ \ Property 3: Intuitive Insight}

A powerful model should not only have high performance but also be interpretable. To the best of our knowledge, only a few approaches propose attention mechanisms to highlight some of the detailed steps, e.g., the representation of the graph and the junction tree. Unfortunately, the principle of optimization, including locating mutated atoms, choosing preserved areas, selecting removed domains and generating extended subgraphs, is discarded.

\section{Experiments}

\subsection{Tasks}
To evaluate the effectiveness, efficiency and explainability of our proposed T{\&}S polish, we apply two typical tasks to simulate real laboratory scenes as in \cite{jin2018learning}:

\textbf{Property Optimization} The novel molecules are generated and optimized according to specified molecular properties, such as solubility or synthetic accessibility. This is a typical application in material science and drug discovery, in which molecules with highly optimized properties may provide a crucial clue for follow-up research.

\textbf{Property Targeting} A property score interval is set for this task, while the aforementioned property optimization task has no property range restriction. In many applications, neither too high nor too low a property score is suitable for a potential candidate molecule.

\subsection{Data}
\begin{table}[!htbp]
  \label{dataset_overview}
  \caption{Basic statistics of the datasets}
  \renewcommand{\arraystretch}{1.3}
  \begin{tabular}{c|ccccccccccc}
    \hline
    Dataset & \#Training & \#Valid & \#Test & $\lambda_1$ & $\lambda_2$ & $\lambda_3$\\
    \hline
    LogP4 & 99,909 & 200 & 800 & 0.4 & - & -\\
    LogP6 & 75,284 & 200 & 800 & 0.6 & - & -\\
    QED & 88,306 & 360 & 800 & 0.4 & [0.7, 0.8] & [0.9, 1.0]\\
    DRD2 & 34,404 & 500 & 1000 & 0.4 & [0, 0.05] & [0.5, 1.0]\\
    \hline
  \end{tabular}
  \begin{tablenotes}
  \item[1] \textbf{note:} The symbol ``-'' indicates that there is no restriction on the property range.
  \end{tablenotes}
\end{table}

Following prior works \cite{you2018graph, jin2018learning, jin2018junction, fu2019core}, we conduct molecular optimization experiments on the ZINC molecule dataset, which consists of 250K drug molecules \cite{sterling2015zinc}. We use the same molecule pairs and train/valid/test split as \cite{jin2018learning}, in which the source and target molecular pairs $(X, Y)$ used for training meet two constraints. First, the similarity between the source molecule $X$ and target compound $Y$ must be above $\lambda_1$. The similarity is calculated according to the Tanimoto distance of the Morgan fingerprints between the two molecules \cite{rogers2010extended}. In addition, compared with $X$, the property value of $Y$ should be improved by an amount within a certain range, i.e., from $\lambda_2$ to $\lambda_3$. The first constraint ensures the \textbf{similarity} of the source and target molecules, while the second constraint guarantees the \textbf{novelty} of the target molecules. In line with different optimization targets, the conditions are assigned different values. We list all these details in Table \ref{dataset_overview}.

Three molecular optimization targets are employed on four different datasets:
\begin{itemize}
  \item \textbf{Penalized LogP}
  The target is intended to improve the LogP score of a molecule in a way that considers the solubility and synthetic accessibility of a given compound \cite{Ertl2009Estimation}. This optimization indicator $\lambda_2$ (or $\lambda_3$) has no bound. In terms of two different similarity constraints $\lambda_1$, two separate datasets are constructed, i.e., LogP4 for 0.4 and LogP6 for 0.6.
  \item \textbf{Drug Likeness (QED)}
  This measurement is used to quantify the probability that a molecule belongs to a drug, which has a fixed range of $[0, 1]$ by definition \cite{bickerton2012quantifying}. The goal of this task is to generate a molecule with a high probability of being a drug.
  \item \textbf{Dopamine Receptor (DRD2)}
  Improving a molecular biological activity against a biological target, the dopamine type 2 receptor, is another goal. The available optimization interval of this property is $[0, 1]$.
\end{itemize}

\subsection{Baselines}
We compare our approach with the following state-of-the-art baselines:
\begin{itemize}
  \item \textbf{GCPN}
  This is a typical reinforcement learning--based approach \cite{you2018graph}. A molecule is generated through iterative addition or removal operations on atoms and bonds in graph convolutional policy networks.
  \item \textbf{JTVAE}
  This is a deep generative model in which latent spaces are learned to generate novel molecules \cite{jin2018junction}. Moreover, scaffolding trees are introduced in the original molecular graphs at both the encoding and decoding stages.
  \item \textbf{VJTNN(GVJTNN)}
  Jin et al. proposed a graph-to-graph translation method to learn a mapping from one molecule to another with better properties \cite{jin2018learning}. To further enforce the naturalness of the generated compound, an adversarial scaffold regularization component is imposed on the VJTNN, and the result is called the GVJTNN.
  \item \textbf{CORE}
  A copy{\&}refine strategy is proposed to determine whether to copy an existent substructure of the input molecule or generate a new component at each prediction step \cite{fu2019core}.
\end{itemize}

\subsection{Implementation Details}
The implementations and hyperparameters of the VJTNN (GVJTNN) \footnote{\url{https://github.com/wengong-jin/iclr19-graph2graph
}}, GCPN \footnote{\url{https://github.com/bowenliu16/rl_graph_generation
}}, JTVAE \footnote{\url{https://github.com/wengong-jin/icml18-jtnn}}, and CORE \footnote{\url{https://github.com/futianfan/CORE}} are completely derived from their original works. For our method T\&S polish, we apply most of the hyperparameters in \cite{jin2018learning}.
Specifically, we train all models with the Adam optimizer \cite{kingma2014adam} for 30 epochs, and the learning rate is $10^{-4}$. In addition, the learning rate is annealed by 0.9 for every epoch in the LogP4 and DRD2 datasets. The batch size is fixed at 32. The dimensions of the hidden state is 300. We execute each generation procedure once and then evaluate the generated molecule.
% The code of the proposed model can be downloaded from \url{https://github.com/aI-area/T-S-polish}.

\subsection{Results}

\subsubsection{Property Targeting}
\begin{table}[!htbp]
  \renewcommand{\arraystretch}{1.3}
  \caption{The definition of three different metrics}
  \centering
  \begin{tabular}{c|c|cc}
    \hline
    \multicolumn{1}{c|}{Metric} & Dataset & $\eta_1$ & $\eta_2$ \\
    \hline
      ~ & QED & [0.3, 1.0] & [0.6, 1.0]\\
      M1 & DRD2 & [0.3, 1.0] & [0.6, 1.0]\\
      ~ & LogP4/LogP6 & [0.4, 1.0] & [0.8, +$\infty$)\\
    \hline
      ~ & QED & [0.4, 1.0] & [0.8, 1.0]\\
      M2 & DRD2 & [0.4, 1.0] & [0.8, 1.0]\\
      ~ & LogP4/LogP6 & [0.4, 1.0] & [1.2, +$\infty$)\\
    \hline
      ~ & QED & [0.4, 1.0] & [0.9, 1.0]\\
      M3 & DRD2 & [0.4, 1.0] & [0.5, 1.0]\\
      ~ & LogP4/LogP6 & [0.4, 1.0] & [0.6, +$\infty$)\\
    \hline
  \end{tabular}
  \label{distinct_metrics}
  \begin{tablenotes}
  \item[1] \textbf{note:} These metrics are defined according to different similarities ($\eta_1$) and target ranges ($\eta_2$).
  \end{tablenotes}
\end{table}

As in the aforementioned definition of the property targeting task, we record the source and generated molecule pair similarity range as $\eta_1$ and the property optimization range as $\eta_2$ to validate performance.
Moreover, as in previous work \cite{jin2018learning, fu2019core}, we apply three distinct measurement metrics for $\eta_1$ and $\eta_2$, as shown in Table \ref{distinct_metrics}.
M2 is stricter than M1, with higher similarity and property targeting. Compared with M2, M3 is a more complex metric, which imposes a higher property targeting on QED and a lower targeting on the DRD2, LogP4 and LogP6 datasets. We examined and counted the generated molecules satisfying the two constraints. The experimental results represent the percentages of qualified target molecules.

\begin{table*}[!htbp]
  \renewcommand{\arraystretch}{1.3}
  \caption{Percentages of successfully generated target molecules in property targeting task}
  \centering
  \begin{tabular}{c|c|ccccc|c}
	  \cline{1-8}
	  \multicolumn{1}{c|}{Dataset} & Metric & GCPN & JTVAE & VJTNN & GVJTNN & CORE & T\&S polish \\
    \hline
      ~ & M1 & 68.25 & 69.03 & 52.75 & 51.38 & 58.54 & \textbf{69.38} \\
      QED & M2 & 33.75 & 37.75 & 25.63 & 25.75 & 32.02 & \textbf{38.38} \\
      ~ & M3 & 8.28 & 9.13 & 17.07 & 17.49 & 18.62 & \textbf{22.53} \\
    \hline
      ~ & M1 &	9.52 & 3.56 & 50.19 & 52.84 & 53.22 & \textbf{54.54} \\
      DRD2 & M2 & 2.09 &	1.34 & 19.26 & 20.59 & 20.83 & \textbf{22.84} \\
      ~ & M3 & 4.92 & 2.66 & 36.14 & 37.27 & 35.89 & \textbf{38.41} \\
    \hline
      ~ & M1 & 29.63 & 38.95 & 36.25 & 36.75 & 40.13 & \textbf{42.25} \\
      LogP4 & M2 & 16.38 & 35.88 & 35.25 & 35.75 & 38.50 & \textbf{38.75} \\
      ~ & M3 & 35.88 & 39.50 & 36.75 & 37.38 & 41.01 & \textbf{43.13} \\
    \hline
      ~ & M1 & 27.13 & 39.04 & 57.75 & 55.04 & 47.25 & \textbf{86.63} \\
      LogP6 & M2 & 14.88 & 35.87 & 52.38 & 49.87 & 44.01 & \textbf{83.37} \\
      ~ & M3 & 32.13 & 45.02 & 59.38 & 54.75 & 49.38 & \textbf{88.54} \\
    \hline
  \end{tabular}
  \label{property_targeting}
\end{table*}

% General description in M1
As shown in Table \ref{property_targeting}, JTVAE performs better on QED, CORE works better on DRD2 and LogP4, and the VJTNN gains the competitive edge on LogP6. However, none of them can obtain the consistently best performance across all the datasets. Comparatively, our proposed T\&S polish method consistently outperforms all baseline methods across all datasets with all three metrics in which $\eta_1$ focuses on similarity and $\eta_2$ emphasizes on novelty.

In addition, even compared with the best method, the VJTNN, our method clearly shows a huge advantage: an improvement of 28.88\%, 30.99\%, and 29.16\% on LogP6 with M1, M2, and M3, respectively.
It is observed that the extracted training molecule pairs in LogP6 have a greater similarity. The required minimal pair similarity defined in LogP6 is 0.6, while that of the other datasets is 0.4. Our polishing paradigm can find more preserved parts of the molecule pairs with a higher similarity between the source and target molecules. We will give a deeper analysis of this later.
% Account for low performance in JTVAE and GCPN
The performances of GCPN and JTVAE are relatively poor, which is consistent with the results of \cite{jin2018learning}.

\subsubsection{Property Optimization}
Unlike property targeting tasks, property optimization merely limits the similarity between the input and generated molecules. We record the average improved property value among the molecule pairs satisfying the similarity restriction. We adapt two metrics M4 and M5 and set the similarity threshold to 0.3 for M4 and 0.4 for M5. Clearly, M5 calls for a higher similarity than M4.

\begin{table*}[!htbp]
  \renewcommand{\arraystretch}{1.3}
  \caption{Performance on the property optimization task under the M4 and M5 metrics}
  \centering
  \begin{tabular}{c|cccc|cccc}
    \hline
    \multicolumn{1}{c|}{}& \multicolumn{4}{c|}{M4} & \multicolumn{4}{c}{M5}\\
    \multicolumn{1}{c|}{Method} & QED & DRD2 & LogP4 & LogP6 & QED & DRD2 & LogP4 & LogP6 \\
    \hline
    GCPN & 0.06 & 0.12 & 0.82 & 0.81 & 0.04 & 0.05 & 0.50 & 0.45 \\
    JTVAE & 0.06 & 0.05 & 1.69 & 1.66 & 0.05 & 0.03 & 1.08 & 1.17 \\
    VJTNN & 0.07 & 0.52 & 2.37 & 1.66 & 0.03 & 0.28 & 1.27 & 1.45 \\
    GVJTNN & 0.07 & 0.52 & 2.10 & 1.44 & 0.04 & 0.30 & 1.19 & 1.25 \\
    CORE & 0.08 & 0.52 & 2.26 & 1.48 & 0.05 & 0.29 & 1.31 & 1.17 \\
    \hline
    T{\&}S polish & \textbf{0.09} & \textbf{0.54} & \textbf{2.60} & \textbf{2.13} & \textbf{0.06} & \textbf{0.32} & \textbf{1.32} & \textbf{2.02} \\
    \hline
  \end{tabular}
  \label{property_optimization}
\end{table*}

The results for these metrics are shown in Table \ref{property_optimization}. The proposed method achieves the best performance on all the tasks.
Compared to LogP4 and LogP6, in both the QED and DRD2 datasets, the improvement value is relatively small, always below 0.1. We recall that QED qualifies the probability that a molecule belongs to a drug and DRD2 records a molecular biological activity against the dopamine type 2 receptor. The available property ranges for these two optimization targets are projected into $[0, 1]$. The penalized LogP score has an unbounded range. Consequently, the improved values on QED and DRD2 are much smaller than that on LogP4 and LogP6.

\subsection{Extensive Study}

\subsubsection{Effect of the Graph Polish Paradigm}

\label{Effect_of_Graph_Polishing_Paradigm}

In this section, we deeply investigate the reason why our method can perform well in all the conducted experiments.
%Considering that the trends in both the tasks and metrics are similar, we choose the trained models in M3 as our study subjects.
Considering that M3 is a more complex metric than M1 and M2, which imposes a higher property targeting on QED and a lower targeting on the DRD2, LogP4 and LogP6 datasets, we choose the trained models in M3 as our study subjects.

Recalling the core motivation of our work, we consider the molecular optimization task as a problem of graph polishing. According to this paradigm, the T\&S polish framework aims to guarantee the maximal preservation region in the source compounds and minimize the removed branches. Benefitting from this mechanism, the scales of the necessarily added branches are tremendously reduced, which provides the potential for T\&S polish to effectively generate the remaining limited but effective subgraphs. We now give the detailed distribution of these branches.

We carry out molecule generation with the trained models on the test sets of QED, DRD2, LogP4 and LogP6. Then, we quantify the scales of the preservation, removal and addition subgraphs for each input molecule. Specifically, we define the scale of a subgraph as the number of atoms in this region. Finally, the three corresponding average numbers of atoms among all input molecules are computed.

\textbf{$\bullet$ \ Why can our method effectively outperform others?}

\begin{figure}
 \centering
 \includegraphics[width=0.49\textwidth]{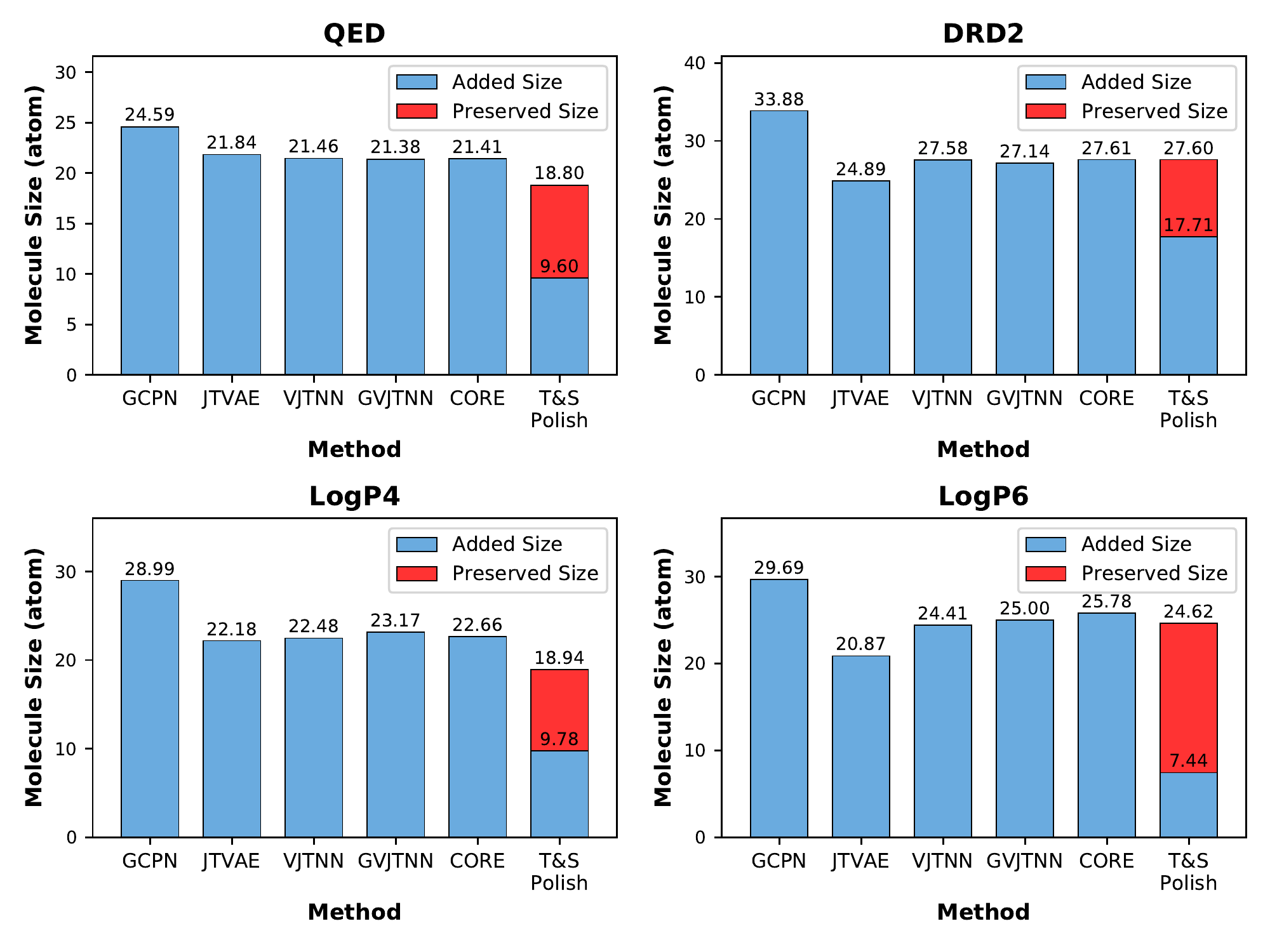}
 \caption{Distribution of the scales of the generated subgraphs across all methods.}
 \label{add_branch_distribution}
\end{figure}
This question can be answered with Fig. \ref{add_branch_distribution}. Other than the only reinforcement learning method, GCPN, we observe that the numbers of atoms in molecules generated by the baselines and T\&S polish are almost equal. That is, these methods can all be expected to generate molecules on a similar scale.
Due to the large set of possible substructures in generating additional branches, most errors occur in this stage. Accordingly, the generation process in T\&S polish is designed to be composed of two steps: preserving existing branches and generating additional branches. T\&S polish avoids this issue as much as possible and only needs to generate limited subgraphs, with scales of 9.60, 17.71, 9.78 and 7.44 in QED, DRD2, LogP4 and LogP6, respectively. For comparison, the average scales for the baseline methods sharply increase to 22.14, 28.22, 23.90 and 25.15.
The fewer newly generated substructures there are, the less chance of making mistakes.

\textbf{$\bullet$ \ Why is the advantage of T\&S polish on the LogP6 dataset so great?}
\begin{figure}
 \centering
 \includegraphics[width=0.49\textwidth]{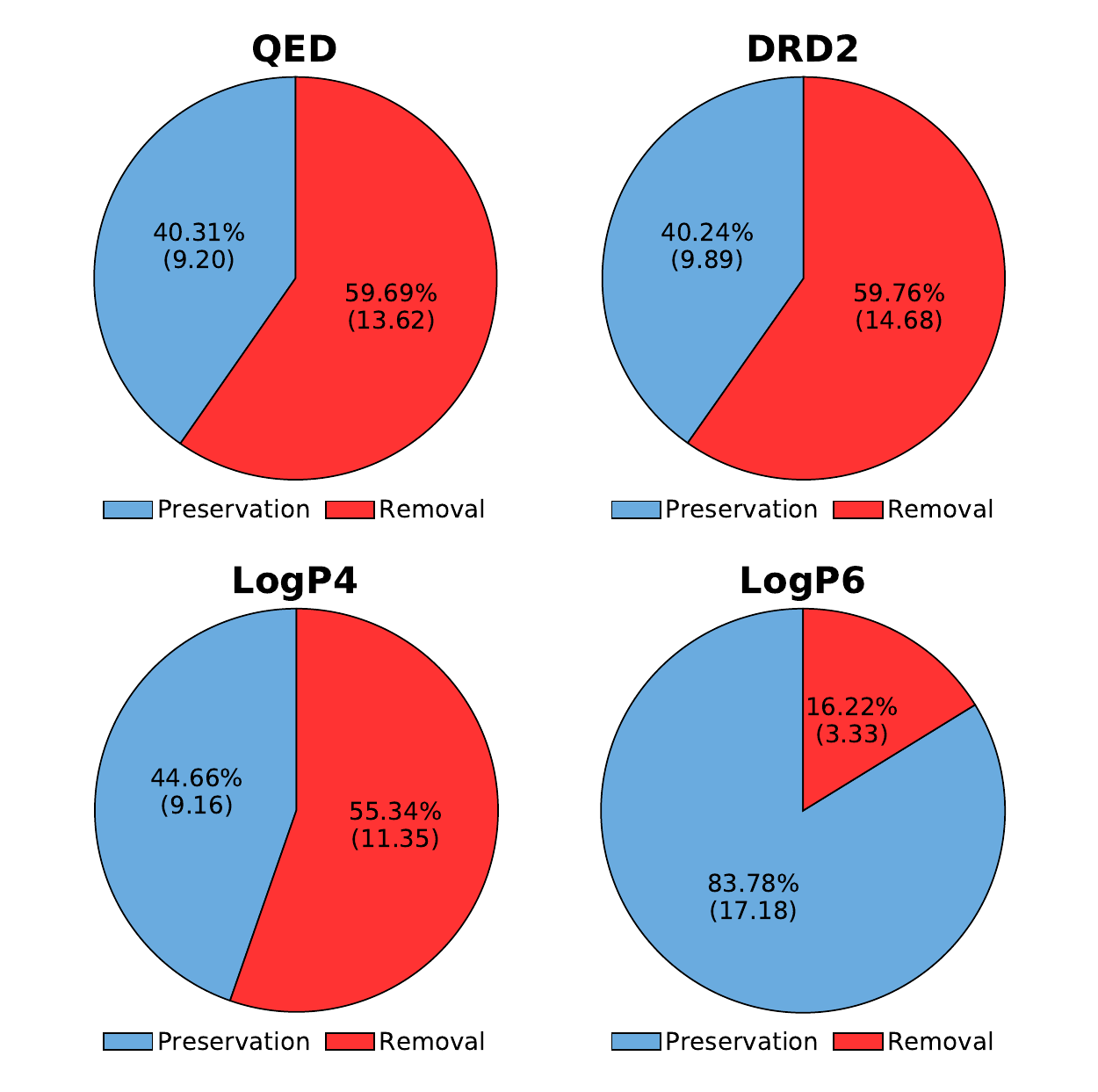}
 \caption{Comparison of the scale of the preservation and removal subgraphs selected by T\&S polish. The bracketed numbers indicate the average numbers of atoms in molecules.}
 \label{branch_distribution}
\end{figure}

Both Table \ref{property_targeting} and \ref{property_optimization} show the tremendous advantage of T\&S polish on the LogP6 dataset. We can gain insight into this with the distribution of the scales of the preservation and addition areas. Fig. \ref{branch_distribution} shows that the preservation subgraphs in QED, DRD2 and LogP4 are slightly smaller than the removal area, which reasonably reduces the load of the later generation process.
Nevertheless, in LogP6, the preservation branches are much larger than the removal branches. More importantly, this large preservation scale directly results in significantly smaller addition subgraphs, i.e., 7.44 in Fig. \ref{add_branch_distribution}.
The reason for this is that the similarity limitation on LogP6 is 0.6 instead of 0.4 which is applied in the other datasets.
This observation is unambiguously consistent with the answer to the first question, that having fewer additional branches can significantly improve the experimental performance.

\subsubsection{Explainability of the Graph Polish Paradigm}

Many graph generators are designed as powerful black boxes, in which performance is achieved but transparency is neglected. Fortunately, in our proposed method, both performance and transparency are guaranteed.
The molecule generation procedure is decomposed into three components: predicting the optimization center, determining the preserved and removed branches, and generating the added branches. These successive steps shed light on the ``logic'' of our model. In this section, we give some examples of the inner step-by-step decision process of our trained model and demonstrate how to understand why our model gives these responses.

The built-in RDKit function uses the method of \cite{bickerton2012quantifying} to calculate the QED score, in which eight widely used molecular properties are considered vital clues. One of these properties is the number of alert structures in the molecule. These alert structures, defined in \cite{brenk2008lessons}, are certain unwanted functionalities in drug-like molecules that give us a direct way to determine which parts of the molecule might have a negative contribution to the QED score. Molecules with more alert structures could have lower QED scores. Therefore, in this experiment, we randomly choose some source molecules with alert structures in the QED test set and feed them into our trained model. Then, we observe how the source molecules evolve in our T\&S polish method.

\begin{figure*}
 \centering
 \includegraphics[width=0.98\textwidth]{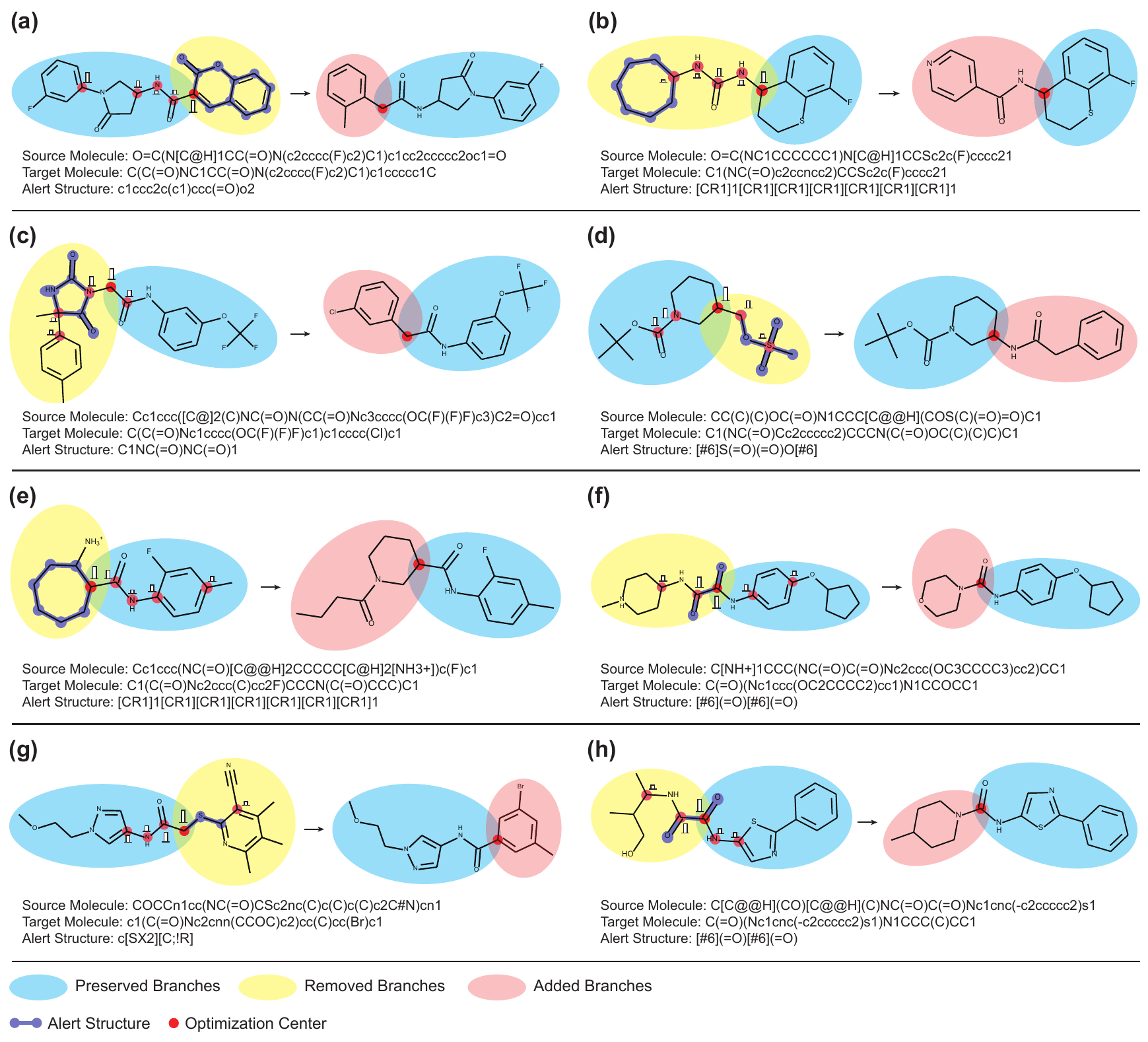}
 \caption{Some successful generation pairs illustrating how our model can generate the required molecules. The corresponding SMILES of the molecules and the SMARTS \cite{james1995daylight} of the alert structures are provided below the 2D structures of the pairs. The normalized probabilities of the top 5 ranked atoms, to be considered optimization centers, are drawn as bars.
It is common for T\&S polish to delete branches with alert structures and preserve other suitable branches. In most cases, the optimization centers is located near the alert structures so that our method can delete unwanted structures with less loss of ``innocent'' substructures.}
 \label{QED_pairs_with_alert_eliminate}
\end{figure*}

Fig. \ref{QED_pairs_with_alert_eliminate} shows some examples of successfully generated molecules under the M3 metric. T\&S polish always chooses the atoms near the alert structures as optimization centers, cleverly eliminates these unsuitable alert structures, and selectively preserves other branches which may make positive contributions to the current optimization task.
We also provide a quantitative analysis of the extent of the decisiveness with which T\&S polish makes the correct decision. We show the top 5 normalized probabilities of the atoms to be considered as optimization centers, $s^{st}_i$ in Equation (\ref{Method:Predicting Optimization center:E5}), as bars in Fig. \ref{QED_pairs_with_alert_eliminate}. It is observed that T\&S polish has the ability to choose the correct optimization centers from candidate atoms with less hesitation. This is a logical chain that is a reasonable way to generate a molecule with a high QED score. From another perspective, we can also decide whether the responses given by T\&S polish are rational and can even explore the potential rules for generating novel molecules in later researches.

\subsubsection{Efficiency of the Graph Polish Paradigm}

In this section, we further investigate the training cost of each method. We take the datasets LogP4 and LogP6, and metric M4 as our main subjects and JTNN, GVJTNN and CORE as the references. The source codes of these methods can be obtained from their corresponding published websites, and they all support GPU acceleration techniques. Our server environment consists of a 3.00 GHz Intel Xeon CPU E5-2687W, 512 GB memory and an Nvidia Titan RTX GPU.

\begin{figure}
 \centering
 \includegraphics[width=0.49\textwidth]{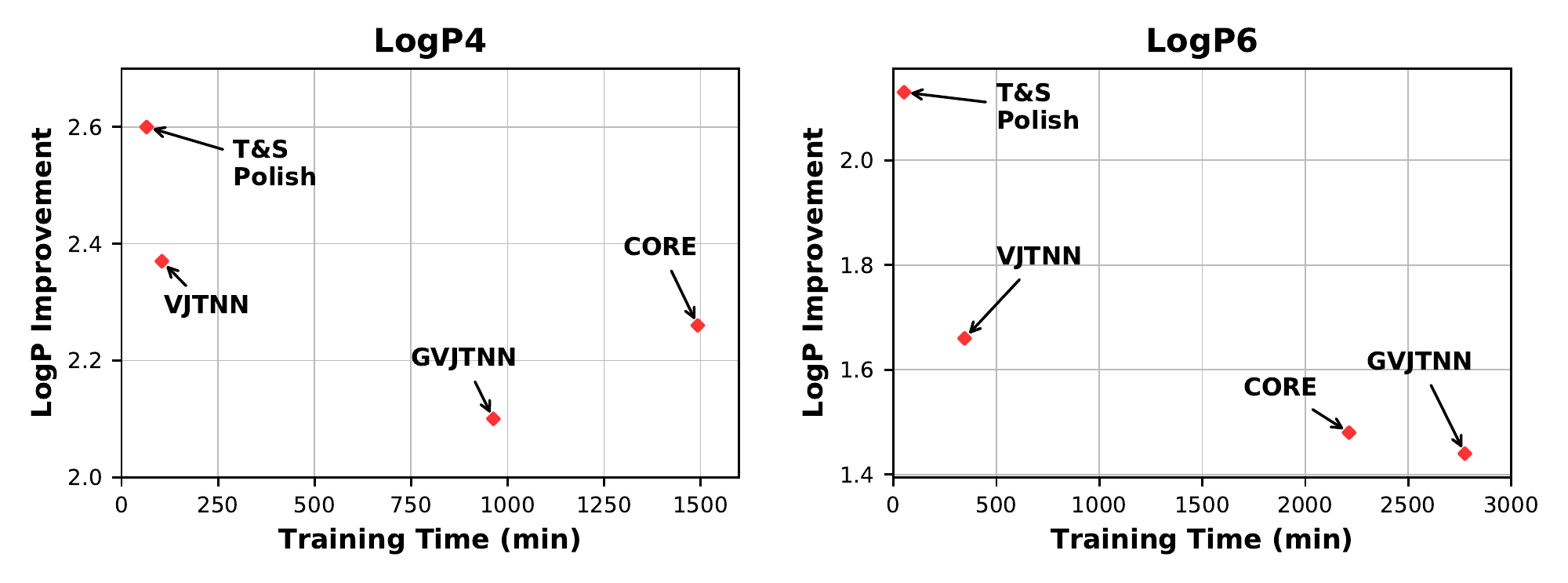}
 \caption{Comparison of the training time and performance. The Y-axis records the performance of the VJTNN, the GVJTNN, CORE and T\&S polish in property optimization tasks under metric M4. The X-axis corresponds to the total training time.}
 \label{training_time}
\end{figure}

The detailed training time curves are shown in Fig. \ref{training_time}. The training time for each epoch in different methods differs greatly. In the LogP6 dataset, CORE consumes the maximum training time, 316 minutes/epoch, followed by 278 minutes/epoch for GVJTNN. Comparatively, VJTNN costs less time, i.e., 43 minutes/epoch. Furthermore, T\&S polish only needs 21 minutes for a single epoch.

In addition to the training time of the S component, we also recorded the computational time taken by the T component: approximate 10 and 9 minutes for LogP4 and LogP6, respectively.
Adding the training time of each epoch in the S component, and the extra time used by the T component, the total time cost of T\&S polish accounts for approximately 2.3\% of CORE and 14.8\% of the VJTNN. Furthermore, the overall tendency of the computational time in the LogP4 dataset is similar as that in LogP6.

\section{Conclusion and Future Work}
This paper proposes a novel molecule optimization paradigm, Graph Polish. Unlike the current translation paradigm, which generates a target molecule by adding substructures one by one from scratch, the polishing paradigm automatically generates a relatively small novel part of the target molecule by maximizing the preserved parts of the source molecule. In this way, the preserved areas greatly decrease the number of steps of molecular optimization and offer an effective clue for the subsequent generation of the new substructures as a prior knowledge.
Furthermore, an effective and efficient learning framework, T\&S polish, composed of T\&S components is proposed to capture the long-term dependencies in the optimization steps. The T component automatically identifies the optimization center and the preservation, removal and addition of certain parts of the molecule, while the S component learns these behaviors and leverages the actions to construct a new molecule.
An intuitive interpretation for each molecular optimization output is naturally produced by the proposed T\&S polish approach.
Experiments with multiple optimization tasks on four benchmark datasets show that the proposed T\&S polish approach significantly outperforms the state-of-the-art baseline methods. Extensive studies are conducted to validate the effectiveness, explainability and time savings of the novel optimization paradigm.

\ifCLASSOPTIONcaptionsoff
  \newpage
\fi



% \bibliographystyle{IEEEtran}
% \bibliography{bare_jrnl}
\bibliographystyle{IEEEtran}
\bibliography{main}
\end{document}